%% file: root.tex
\documentclass[letterpaper, 10 pt, conference]{ieeeconf}  

\usepackage{tabularx} 
\usepackage{graphicx}
\usepackage{mathtools}
\usepackage{bm}
\usepackage{epstopdf}

\usepackage{hyperref}

\hypersetup{
	bookmarks=true,         
	unicode=false,          
	pdftoolbar=true,        
	pdfmenubar=true,        
	pdffitwindow=false,     
	pdftitle={ESD CYCLOPS: A new robotic surgical system for GI surgery},    
	pdfauthor={T.J.C. Oude Vrielink, Ming Zhao, Ara Darzi and George P. Mylonas},     
	pdfsubject={Surgical Robotics},   
	pdfcreator={T.J.C. Oude Vrielink},   
	pdfproducer={},  
	pdfkeywords={CYCLOPS} {Surgical Robotics} {Cable-Driven Parallel Mechanism}, 
	pdfnewwindow=true,      
	colorlinks=false,       
	linkcolor=red,          
	citecolor=green,        
	filecolor=magenta,      
	urlcolor=cyan           
}

\IEEEoverridecommandlockouts                              

\title{\LARGE \bf
	ESD CYCLOPS: A new robotic surgical system for GI surgery
}

\author{Timo J.C. Oude Vrielink, Ming Zhao, Ara Darzi and George P. Mylonas, \textit{Member, IEEE} 
\thanks{HARMS Lab, Department of Surgery, Imperial College London, W2 1PF London, UK
        {\tt\small {t.oude-vrielink15}@imperial.ac.uk}}%
}

\begin{document}

\maketitle
\thispagestyle{empty}
\pagestyle{empty}

\begin{abstract}
\input{./body/Abstract.tex}

\end{abstract}

\section{INTRODUCTION}
	\input{./body/Introduction.tex}


\section{CYCLOPS Robotic System Design}
	\input{./body/SystemOverview.tex}


\section{Deployable Scaffold}
	\input{./body/ScafIntro.tex}

	\subsection{Parametric Scaffold Design}
		\input{./body/ScafParam.tex}

	\subsection{Clinical Design Criteria}
		\input{./body/ClinicalCriteria.tex}

	\subsection{Scaffold Embodiment}
		\input{./body/ScafEmbod.tex}

	\subsection{Scaffold Deployment Actuation}
		\input{./body/actScaffold.tex}

\section{System Validation}
 The system has undergone scrutinisation through technical and pre-clinical validation. Initially, the accuracy and force exertion capabilities of the system have been assessed through bench tests, discussed in Section IV.A. In addition, the system has been subjected to \textit{ex vivo} and \textit{in vivo} validation, as discussed in Section IV.B. 
 
	\subsection{Bench Testing}
		\input{./body/benchtop}

	\subsection{Pre-clinical validation} 
	
		\begin{figure}[h]
			\centering
			\includegraphics[width=3.3in]{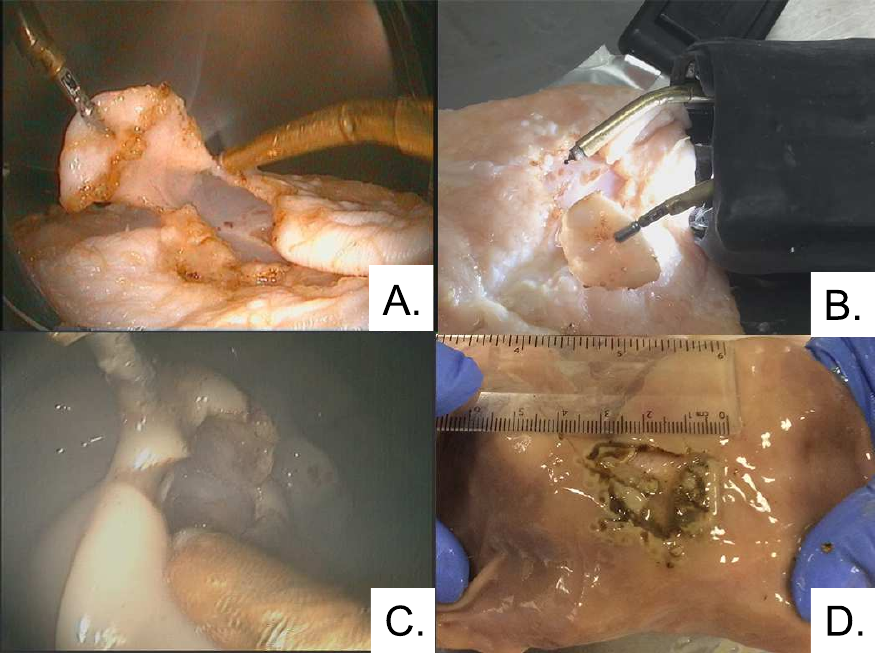}
			\caption{Images from the Ex Vivo validation. \textbf{A.} Endoscopic view while dissecting skin from chicken thigh. \textbf{B.} Exterior view showing the dissected lesion (chicken thigh). \textbf{C.} ESD performed in a procine stomach. \textbf{D.} Assessing of the dissected area.}
			\label{fig_EXtrials}
		\end{figure}
\begin{figure}[h]
				\centering
				\includegraphics[width=3.3in]{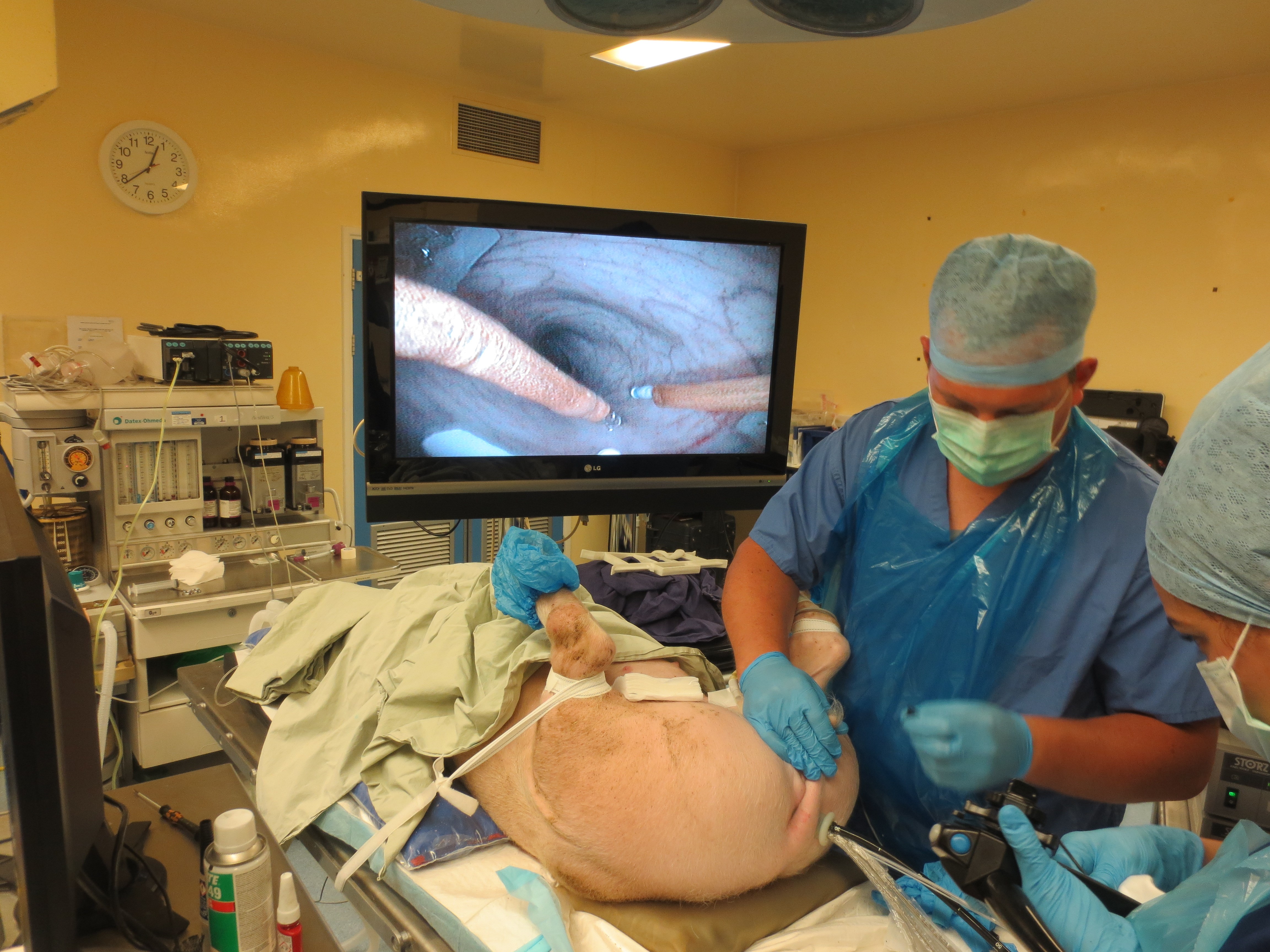}
				\caption{The system is currently undergoing \textit{in vivo} pre-clinical validation on pigs.}
				\label{fig_trials}
			\end{figure}

		To assess the ability to perform ESD, pre-clinical trials were performed by a GI surgeon. \textit{Ex vivo} trials have been conducted on chicken thigh, by dissecting the chicken skin from the flesh (Fig. \ref{fig_EXtrials}.a -b). After success, the deployment in combination with ESD has been performed in a \textit{ex vivo} porcine stomach (Fig. \ref{fig_EXtrials}.c-d). The dissected lesion size corresponds approximately with the 26x30mm workspace discussed in section IIIC. Currently, the system is undergoing \textit{in vivo} pre-clinical validation (Fig. \ref{fig_trials}). Detailed discussions on the pre-clinical data is not the focus of this technical paper, and will shared in later work.

\section{Discussion \& Conclusion}
The current paper has shown and discussed extensive validation of the CYCLOPS system, with a focus on ESD. The scaffold is designed according to clinical requirements for ESD and the parametric design can be customised further to account for specific clinical needs. Concurrently, an optimisation algorithm is being developed to optimise the instrument workspace to simulated surgical tasks \cite{customCYCLOPS}. This specific work focuses on one of previously mentioned core benefits of CDPMs, namely, workspace customisability by changing the tendon entry and attachment points. 

The validation experiments have shown that the mechanism can achieve high forces and is able to accurately perform a tracing task. The curved tool has shown lower forces when compared to the straight tool. This is explained by the additional arm generating a moment around the tool centre of rotation. Specifically in the redundant DOF (end-effector roll), the stiffness is dependent on the antagonistic stiffness caused by the tensions in the other DOF. For the curved tool, the forces in the Z-axis are directly related to the antagonistic angular stiffness around the X-axis. At the homing position, the tendons are equally pre-tensioned to 10N, resulting in a considerable high stiffness in the Z-direction. To increase this stiffness further and throughout the workspace, a more advanced control system is currently being implemented using torque control and an end-effector stiffness observation \cite{gigi}, enabling us to actively control the stiffness in any redundant DOF. Such control system has benefits in terms of preventing tendon slackness, and thus loss of controllability (especially close to singularities), and also will enable higher force exertion by the instruments. 

Nevertheless, even with the position control approach used in the validation, the results are phenomenally good, and the current pre-clinical trials show that the technical advantages make it possible to accurately dissect lesions without limitations in terms of instrument dexterity and forces. These are major improvements compared to other robotic systems, and due to the high customisability of the system, these technical benefits seem promising to be transferred to other surgical domains. 

\section*{ACKNOWLEDGMENT}
This work is supported by the ERANDA Rothshild Foundation. The authors would like to thank George Dwyer and Dan Stoyanov for their help on the manufacturing of the deployable scaffold. 

\addtolength{\textheight}{-12cm}   





\bibliographystyle{IEEEtran}
\bibliography{bibESDCYCLOPS}{}




\end{document}

%% file: body/Abstract.tex
Gastrointestinal (GI) cancers account for 1.5 million deaths worldwide. Endoscopic Submucosal Dissection (ESD) is an advanced therapeutic endoscopy technique with superior clinical outcome due to the minimally invasive and \textit{en bloc} removal of tumours. In the western world, ESD is seldom carried out, due to its complex and challenging nature. Various surgical systems are being developed to make this therapy accessible, however, these solutions have shown limited operational workspace, dexterity, or low force exertion capabilities. The current paper shows the ESD CYCLOPS system, a bimanual surgical robotic attachment that can be mounted at the end of any flexible endoscope. The system is able to achieve forces of up to 46N, and showed a mean error of 0.217mm during an elliptical tracing task. The workspace and instrument dexterity is shown by pre-clinical \textit{ex vivo} trials, in which ESD is succesfully performed by a GI surgeon. The system is currently undergoing pre-clinical \textit{in vivo} validation. 

%% file: body/Introduction.tex
Gastrointestinal(GI) cancers accounted for nearly 40,000 deaths in the UK and 1.5 million worldwide in 2014 \cite{CRUKstats}. Research suggests that if diagnosed and appropriately treated at their earliest stage, survival rates are very high especially for colorectal cancer.
 Constant improvements in endoscopic imaging technology in conjunction with national screening programs, are likely to result in a greater number of polyps and early colorectal cancers been detected. Endoscopic Submucosal Dissection (ESD) is an advanced therapeutic endoscopy technique developed as an alternative to conventional open or laparoscopic surgery for removing early cancers or polyps. In recent years ESD has grown rapidly. The technique has extensively been adopted in Eastern Asia for treatment of early gastric cancer due to its excellent results \cite{puli2009successful}. In the Western world ESD is seldom carried out, especially for the removal of colorectal cancers, due to its complex and challenging nature, longer procedure times of up to four hours and high perforation (up to 11.8\%) and bleeding rates (up to 3\%) \cite{kantsevoy2010w1447}\cite{saito2010prospective}. The complication rates are not insignificant and so there is a need for instrument platforms and tools that are easier and safer to use by reducing the incidence of defects caused by the procedure or allow their reliable closure, should they occur.

\begin{figure}
	\centering
	\includegraphics[width=3.4in]{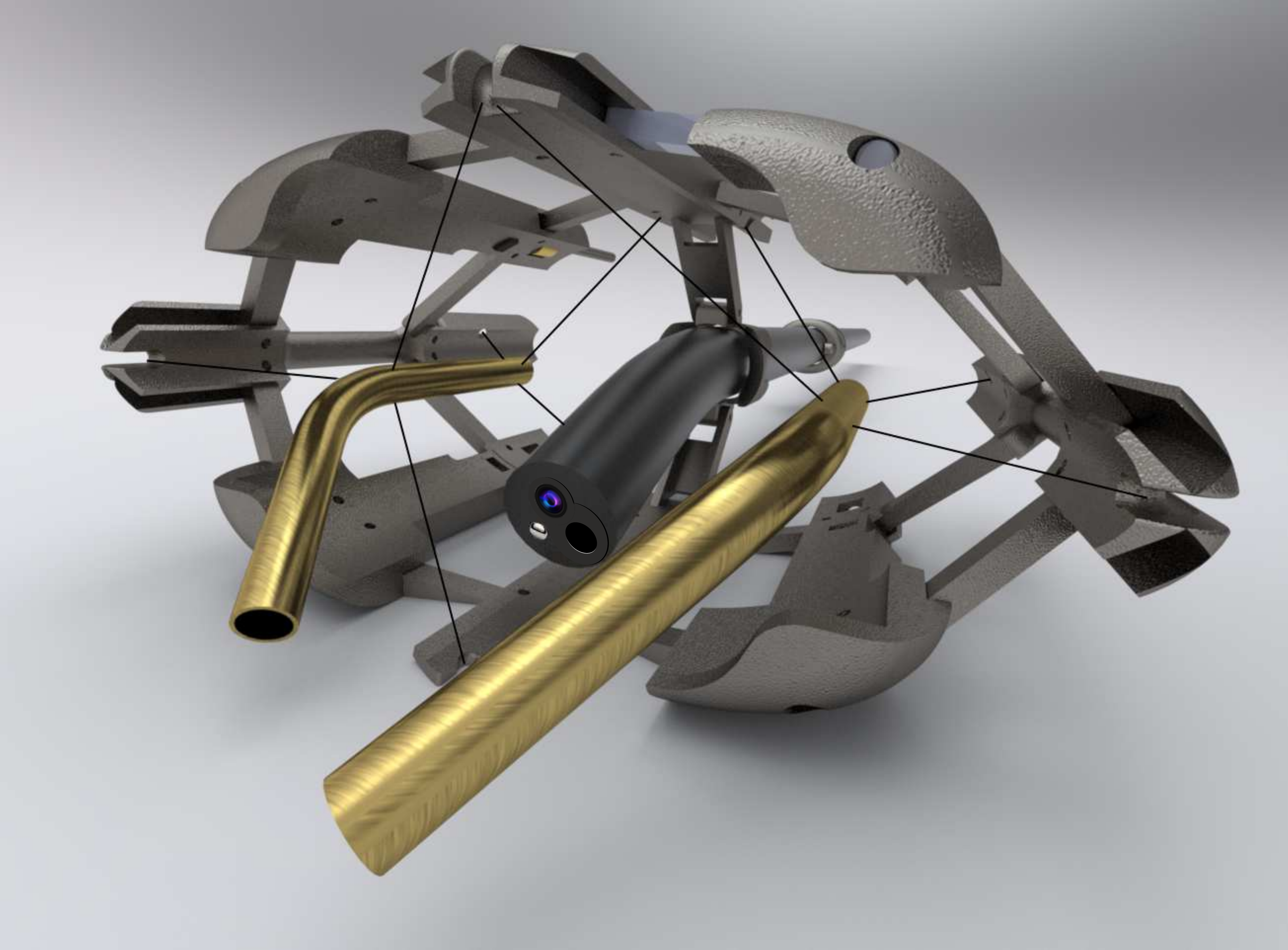}
	\caption{Render of the ESD CYCLOPS system, without the soft silicone sleeve.}
	\label{fig_oldProto}
\end{figure}

Typically, ESD involves an electrosurgical cutting tool introduced via the biopsy channel of a flexible endoscope. The aim is to purposely dissect the submucosa, which is the tissue layer of the GI tract that supports the mucous membrane. Only very limited control of the cutting tool is possible by inserting and withdrawing it from the endoscope's working channel and by steering the endoscope tip using its control dials, which greatly hamper manipulation and fine control of the distal tip. Additionally, during dissection the mucosa above may hang down over the working field, effectively blinding the endoscopist. Furthermore, care should be taken so that the cutting device is operated in angles close to parallel to the GI wall and by using traction movements for avoiding bleeding and perforation. The lack of bimanual dexterity and tissue retraction -commonly referred to as tissue triangulation- are the main contributing factors to the technical complexity and lack of wider adoption of ESD.

To overcome the above limitations, various traction systems and techniques have been described in the literature. However, none of these systems have gained widespread popularity and proof of their efficacy remains scarce \cite{teoh2013ex}. A number of non-commercial robotic systems with \textit{in vivo} ESD trials have also been proposed \cite{phee2012robot}\cite{diana2013endoluminal}. These systems have demonstrated various degrees of improvement over certain aspects of conventional ESD approaches. However, they represent completely new and fairly complex designs not compatible with existing theatre infrastructure and they provide either limited manoeuvrability and degrees of freedom (DOF) or limited operational workspaces and force exertion capabilities \cite{vitiello2013emerging}. Additionally, due to their mechanical complexity and reusable nature, they are expected to have relatively high manufacturing, maintenance and running costs.

The CYCLOPS concept was first introduced in \cite{mylonas2014cyclops} as a universal endoscopic attachment that allows augmentation and re-purposing of any available endoscope by turning it into a frugal surgical robot with bimanual dexterity. The core component of the design is the use of a cable-driven parallel manipulator (CDPM) to manipulate off-the-shelf flexible instruments (e.g., electrocautery, grasper) within a rigid peripheral scaffold. CDPMs have been explored in literature for many different applications, ranging from radiotelescopes, industrial weight lifting, construction crane solutions, assembly, manufacturing, stadium camera's, environment sensing for terrestial and aquatic applications, haptic devices, rehabilitation, search and rescue, motion simulators, windtunnel test \cite{gosselin2014cable}. Typical reasons to use this type of mechanism include:

\begin{itemize}[\setlabelwidth{Z}]
	\item[-] High payload and force transmission 
	\item[-] Large workspace 
	\item[-] Workspace reconfigurability 
	\item[-] High dynamic capabilities
	\item[-] Rapid deployability
	\item[-] Low Costs
\end{itemize}

These properties offer critical advantages in surgery, including bimanual dexterity, improved tissue manipulation and triangulation, high force delivery, large workspace, stability and controllability. These have been documented in \cite{mylonas2014cyclops}, were for instance typical 65N of exerted force capabilities have been demonstrated. In addition, it was shown that the workspace could be optimised according to specific surgical needs, by simply changing the entry points of the tendons into the outer scaffold.

At the early stages of development and for evaluating the proposed concept, the walls of the supporting scaffold were made from solid plastic. This of course prohibits introduction of the endoscope and the CYCLOPS attachment through a natural orifice or small incision. For the design to be practical, the attached unit has to be initially packaged in a very tight space that will allow its introduction inside the body alongside the carrier endoscope. At the same time, the packaged attachment must not interfere with the main functionalities of the endoscope (i.e., flexion, and vision) while it is navigated to the operating anatomy. Once the endoscope is in place, the unit can be deployed in a gradual and controlled manner to form a rigid or semi-rigid scaffold. 

In this paper, the current state of the CYCLOPS system for advanced therapy in GI surgery is discussed. In contrast to previous proof of concept work, the current system takes clinical constraints into account, including the implementation of a deployable scaffold. Accuracy and force exertion capability of the system have been tested via bench testing. Additional validation include pre-clinical \textit{ex vivo} assessment of the system's ability to perform ESD. The developed system is currently undergoing \textit{in vivo} pre-clinical trials.

%% file: body/SystemOverview.tex
\begin{figure}
	\centering
	\includegraphics[width=3.4in]{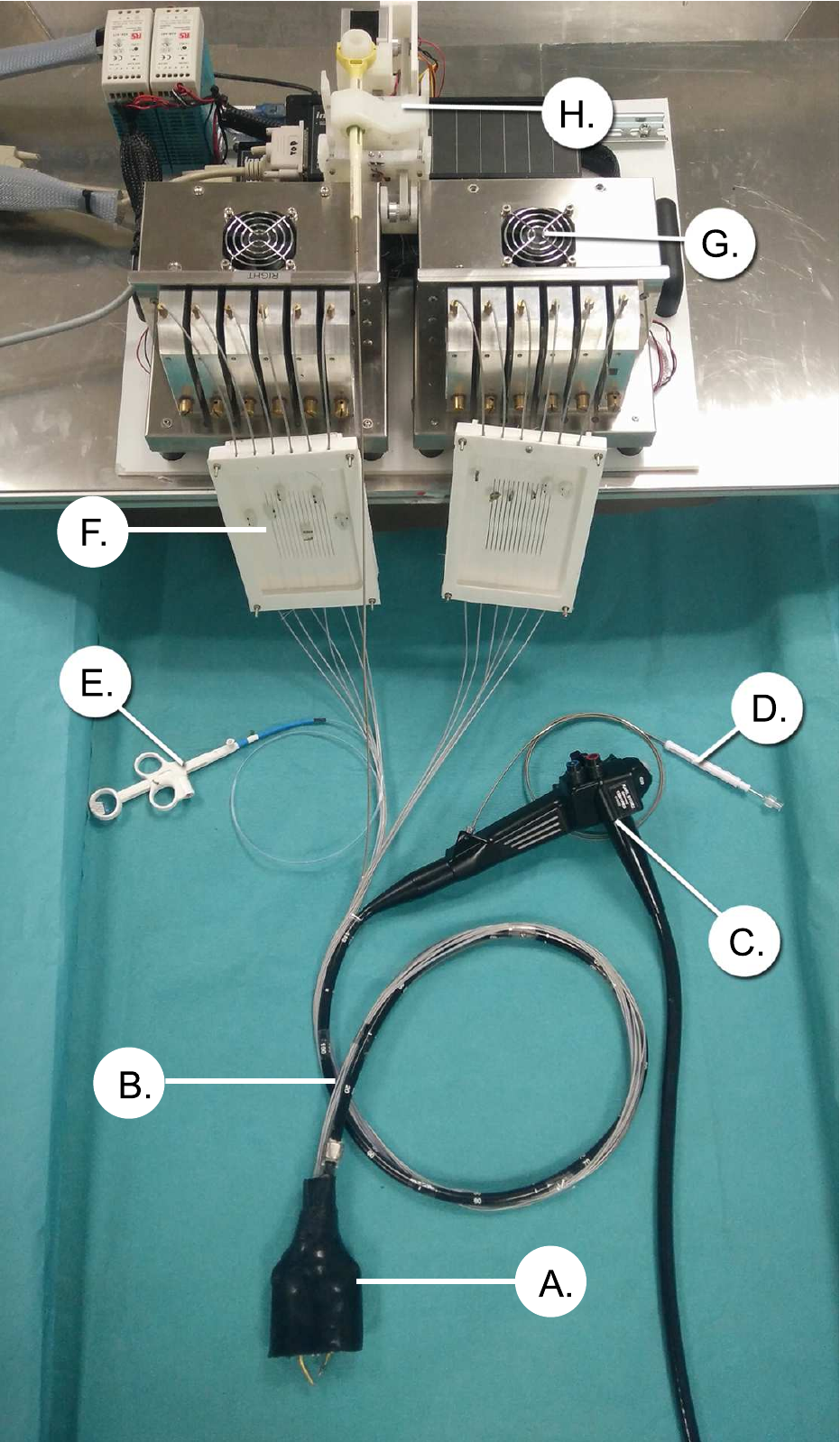}
	\caption{Overview of the mechatronics of the system. \textbf{A.} Deployable scaffold with silicon sleeve and two surgical instruments. \textbf{B.} Bowden cables guiding the tendons from the motor units to the end-effectors.\textbf{C.} Endoscope used for visualisation and as a conduit to place the scaffold in the place of interest. \textbf{D.} The biopsy channel is used for additional instruments. \textbf{E.} Flexible instrument guided to the end-effector. \textbf{F.} Cable-splitters for seperation of the Bowden cables to prevent electrical hazards. \textbf{G.} Motor units for manipulation of the left and right instruments. \textbf{H.} Grasper placed in the actuation mechanism.  }
	\label{fig_systemfoto}
\end{figure}

\begin{figure*}
	\centering
	\includegraphics[width=\textwidth]{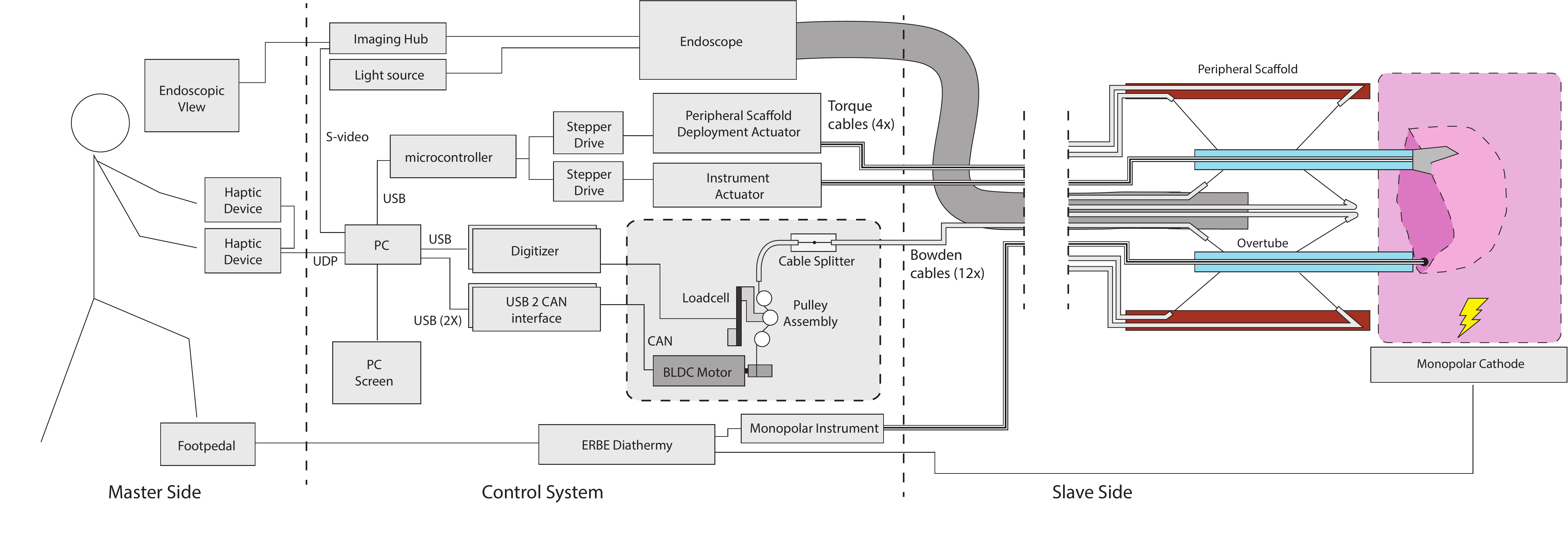}
	\caption{System schematic diagram }
	\label{fig_schematics}
\end{figure*}

The robotic system developed for ESD surgery is shown in Fig \ref{fig_systemfoto}. A schematics describing different elements of the control system is given in Fig. \ref{fig_schematics}. The fully deployed system provides endoscopists with dexterous bimanual control of the introduced instruments through a bimanual master manipulator. 

\subsection{Robotic End-effector} The robotic end-effector - slave side of the system - comprises two instruments that are each controlled using a CDPM. As the cables/tendons can only pull, the number of tendons $n$ must be larger than the number of degrees of freedom (DOF) $m$: $ n \geq m + 1 $. In previous work, it was shown that the  CYCLOPS is able to achieve full 6 DOF \cite{vitiello2014augmented}. The current system implements 6 tendons per instrument, and therefore is controllable in 5 DOF (x,y,z,yaw,pitch).The actuation tendons are connected on one side to overtubes, and are guided from the  scaffold via thin flexible force-transmitting conduits - Bowden cables (1.4mm Round Wire Coil, Asahi Intecc, Japan) - to an external motor unit. Inside the force transmission cables, thin PTFE tubing is placed. Commercially available endoscopic flexible instruments are introduced externally and placed into the overtubes, leaving the biopsy channel(s) of the endoscope available for additional instruments (e.g. flexible needle, additional graspers). 

\begin{figure}[b]
	\centering
	\includegraphics[width=3.3in]{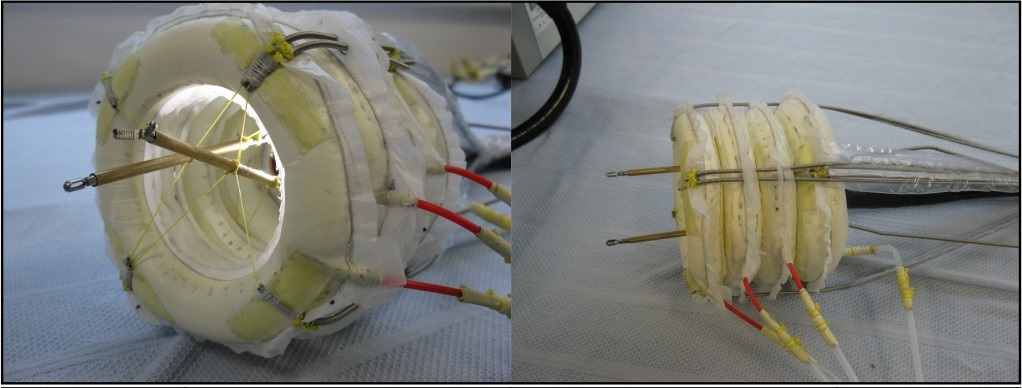}
	\caption{Scaled-up version of the inflatable scaffold.}
	\label{fig_inflatable}
\end{figure}
One of the important aspects of the robotic end-effector is the outer peripheral scaffold. The scaffold has to be deployable to facilitate for minimally invasive access. In previous work, we have developed a scaled-up inflatable version of the scaffold (Fig. \ref{fig_inflatable}). Currently, the mechanism inflatable mechanism is scaled down to the appropriate size for GI endoscopy. Concurrently, we have developed a deployable scaffold using rigid link mechanisms. This is the mechanism discussed in section III of this paper. 

	\subsubsection{Tendon Actuation mechanism}
		\input{./body/actTendons.tex}

	\subsubsection{Actuation of Exchangeable Instruments}
		\input{./body/actInstruments.tex}

\subsection{Master Side} The surgeon manipulates two surgical instruments in 5 DOF via the haptic devices (Geomagic Touch, 3D Systems, USA). The 6th DOF input, the handle roll, is currently not used. Haptic feedback is used to indicate the boundaries of the workspace in which the surgeon can manipulate the instruments. The boundaries given to the surgeon are the singularities of the system. The proximity to a singularity is estimated by use of the \textit{tension factor} \cite{pham2009workspace}, which is the ratio between the maximum and minimum tension in the cables. The minimum and maximum tensions used are theoretical calculations, based on the calculated optimal tension distribution with the L1-norm approach \cite{hay2005optimization}. The buttons on the handles of the haptic devices are used to open/close the grasper, while the second button is used for fixating the handle at a position when the user needs to let off control of the masters. For the diathermy cauthery (VIO 200D, Erbe Elektromedizin GmbH, Germany), the footpedal is used for the different modes of cutting. Visualisation of the scope is provided to the surgeon via a conventional endoscopy imaging system.

%% file: body/actTendons.tex
Each tendon is actuated by direct rotation of a 9mm diameter spool using a brushless DC motors with intergrated motion controller (2232S024BX4 CCD, Faulhaber, Germany). The tension in the cable is measured using a full-bridge strain gauge (LCL-020, OMEGA Engineering, Inc., USA), connected to a DAQ interface (Instrunet i100, GW Instruments, Inc., USA). To prevent any damage to the motors as a result of the electrical surge from the surgical electrocauthery, the steel bowden cables are disconnected from the motor units via a PLA cable-splitter. An additional function of this cable splitter is to facilitate the exchange between different scaffolds. 

%% file: body/actInstruments.tex
The instrument actuator shown in Fig. \ref{fig_systemfoto}H., is designed to facilitate the quick exchange between different flexible surgical instruments. The mechanism is actuated by a stepper motor (103H5208-5240, Sanyo Denki, Japan), moving a timing belt. The instrument holder is mounted on a linear rail, and is coupled to the timing belt. The range of the translational motion is set by limitswitches, which can be positioned manually to accommodate for instruments with a different actuation range.

%% file: body/ScafIntro.tex
The deployable mechanism consists of an inner skeleton and an outer flexible shell. The main function of the inner skeleton is the deployment and rigidity to support exertion of forces by the tendons. The main function of the soft outer shell, made out of silicone rubber (Ecoflex 00-20, Smooth-on, Inc., USA), is protection of tissue during usage of the mechanism and low friction during insertion in the colon. The deployable linkage-based design can be customized according to specific workspace requirements, and therefore, first the general parametric design of this mechanism is discussed.

%% file: body/ScafParam.tex
\begin{figure}
	\centering
	\includegraphics[width=3.4in]{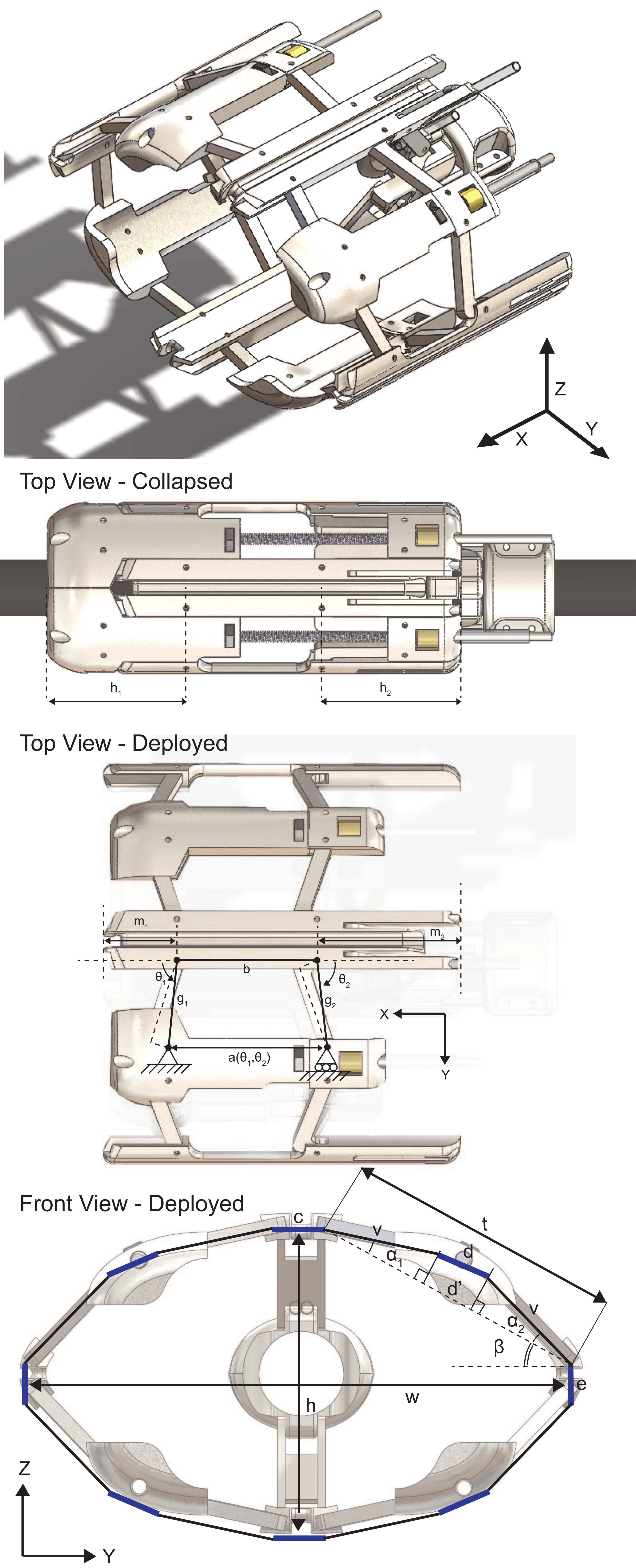}
	\caption{The linkage-based scaffold, supplying the rigidity required for tendon actuation.}
	\label{fig_sim}
\end{figure}

The most fundamental element of the structural design of the scaffold is the 4-bar linkage mechanism, with a single prismatic joint (shown in Fig. \ref{fig_sim}, top View - Deployed). The position of each link of this mechanism is fully determined by angles $\theta_1$ and $\theta_2$, and changing these angles, results in the deployment of the system. 

The deployable scaffold can be defined as two deployable rings. In this section, only the case with symmetry lines along the XY and XZ plane is taken into account. This method can be further extended to non-symmetrical or systems with different symmetry plane.

The size of the deployable ring depends on the projection of the links $ g_1 $ and $ g_2 $ on the YZ-plane (front-view):
\begin{equation}
	v_{i}(\theta_i)=g_i sin(\theta_i)
	\label{eq_first}
\end{equation}

The height and length of each ring - whether deployed or undeployed - are described as:
\begin{equation}
	h=e+ 2t sin⁡(\beta)
	\label{eq_height}
\end{equation}
\begin{equation}
	w=c+ 2t cos⁡(\beta)
	\label{eq_width}
\end{equation}

where
\begin{equation}
	t(v_1,v_2)=d'(v)+v_1cos⁡(\alpha_1 )+v_2cos⁡(\alpha_2 )
\end{equation}
and $ d' $ is the projection of $ d$ on line $ t$: 
\begin{equation}
	d'(v_1,v_2)=\sqrt{d^2-(v_i sin(\alpha_1) - v_2sin(\alpha_2))^2}
\end{equation}

The design parameters are $b$, $c$, $d$, $e$,  $\beta$,  $\alpha_1$, $\alpha_2$, $ g_1 $ and $ g_2 $. $c$, $d$ and $e$ are constants depending on spatial requirements for the joints. The parameters $\alpha_1$ and $\alpha_2$ determine the elliptical shape of the rings. It can be easily seen that $\beta$ determines the ratio between the height and width of the scaffold: $r=h/w$. The maximum and minimum size of the scaffold depends on  $\{v_{min},v_{max} \}$ by:
\begin{equation}
	v_{i,min}=g_i sin(\theta_{i,min})
	\label{eq_vMin}
\end{equation}

\begin{equation}
	v_{i,max}=
	\begin{cases}
	g_i ,& \textit{if } \frac{\pi}{2}\exists [\theta_{i,min},\theta_{i,max}]\\
	g_i sin(\theta_{i,max}), & \textit{otherwise}
	\end{cases}
	\label{eq_last}
\end{equation}

The size of the undeployed scaffold is an important design parameter for minimally invasive surgical applications. The above equations can be used to calculate the undeployed width and height of the mechanism. Also, too long scaffolds will limit the ability of the endoscope to navigate through the colon. The minimal height and width, $ h_{min}$ and $w_{min}$, are found by combining (\ref{eq_vMin}) with (\ref{eq_height}) and (\ref{eq_width}). Visa versa, for the dimensions of the deployed scaffold, (\ref{eq_last}) is used.

The length of the undeployed scaffold depends primarily on whether the mechanism is inward or outward folding upon deployment. For each case, the length of the undeployed mechanism is found as:\\
\begin{equation}
	L_{undeployed,in}=b+h_1+h_2+2g cos⁡(\theta_{min} )
\end{equation}
\begin{equation}
	L_{undeployed,out}=b+m_1+m_2	
\end{equation}
For restrictions in length of the undeployed scaffold, the outward folding mechanism is more beneficial, however, as will become clear later, the choice of deployment direction depends on various factors ranging from actuation method to clinical needs. \\

%% file: body/ClinicalCriteria.tex
For the design of the scaffold, clinical criteria have to be taken into account. Firstly, the deployed design is limited to the colon diameter, which varies strongly depending on the location in the colon. In general, the largest sections are the ascending colon ($61.3\pm 11.1mm$ diameter) and the cecum ($75.7\pm 12.2mm$ diameter), in contrast to the smallest section which is the sigmoid ($34.5\pm 7.1mm$) \cite{khashab2009colorectal}. Slightly smaller diameters have been found during a large-scale Japanese study \cite{sadahiro1992analysis}. A second important clinical aspect is the workspace required for performing ESD. As the entry points of the tendons into the scaffold determine the workspace of the instruments, this should be taken into account into the scaffold design. ESD is the only minimally invasive method to dissect flat and depressed lesions en-bloc that are larger than 20m diameter. One study shows a mean tumour diameter of 20.8mm, when only taking lesions larger than 10mm into account \cite{rembacken2000flat}. A large scale meta-analysis has shown tumour sizes range from 6.2-43.6mm diameter, showing a 32.4mm median value found for the mean tumour size across literature \cite{repici2012efficacy}. 

Other clinical requirements include the access to the site of interest. Based on clinical input and pre-clinical validation, the current size and bullet-shape scaffold have been determined.

%% file: body/ScafEmbod.tex
\begin{table*}[t!]
	\renewcommand{\arraystretch}{1.3}
	\caption{Design parameters}
	\label{table_DesignParam1}
	\centering
	\begin{tabular}{c|c |c |c|c|c|c |c|c |c |c|c}
		\hline
		\bfseries b & \bfseries c & \bfseries d & \bfseries e & \bfseries $g_i$ & \bfseries $h_1$& \bfseries $h_2$ & \bfseries $\theta_{min}$ & \bfseries $\theta_{max}$& \bfseries $\alpha_{1}$& \bfseries $\alpha_{2}$& \bfseries $\beta$\\
		\hline
		\hline
		24mm & 6.5mm & 5.0mm & 4.6mm & 15mm & 25.7mm &25.7mm & 20 deg & 89 deg & 16.5deg & 17.7deg& 28.6deg\\
		\hline
	\end{tabular}
\end{table*}

\begin{figure}[b]
	\centering
	\includegraphics[width=3.5in]{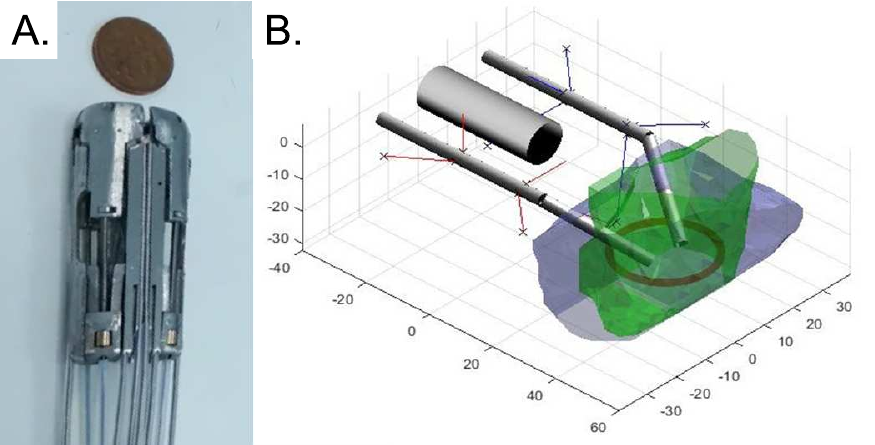}
	\caption{\textbf{A.} The undeployed scaffold. \textbf{B.} Workspace of both instruments, showing a 20mm lesion placed at the bottom of the scaffold. The overlap between both instruments corresponds approximately to a 26x30mm square.}
	\label{fig_WS}
\end{figure}

The current embodiment of the deployable scaffold is created from stainless steel (LaserForm 17-4PH) using Direct Metal Printing (Prox DMP 100, 3D Systems, USA).  One major benefit of this manufacturing method, in particular in combination with CDPMs, is the ability to customise it on a per-case basis. The inward moving sections are accomodated with a bulletshape tip, used to facilate insertion. The design parameters of the specific design are shown in Table \ref{table_DesignParam1}. The dimensions of the scaffold, deployed and undeployed, are given in Table \ref{table_Dimensions}. The undeployed scaffold is shown in Fig. \ref{fig_WS}A.

The deployed circumference corresponds to the 50mm mean diameter found for the transverse colon, and is smaller than the cecum, ascending and rectal sections of the colon anatomy \cite{khashab2009colorectal}.

\begin{table}[h!]
	\renewcommand{\arraystretch}{1.3}
	\caption{Dimensions of the scaffold, Deployed and Undeployed}
	\label{table_Dimensions}
	\centering
	\begin{tabular}{r||c|c}
		\hline
		\bfseries & \bfseries Undeployed & \bfseries Deployed\\
		\hline\hline	
		Width & 30mm & 66mm\\
		Height & 18mm & 39.60mm\\
		Length & 72.9mm & 61mm\\	
		Circumference & 83.2mm & 162.2mm \\
		\hline
	\end{tabular}
\end{table}

As the required taskspace is on the surface of the colon, the overtubes are accomodated with a curvature. The homing position is set to the centre of the workspace, the curvature is defined in such a way that the instruments centre of the workspace is in the centre of the taskspace. The zero-wrench reachable workspace can be calculated using the optimal tension distribution calculated with the L1-norm \cite{hay2005optimization}. The results are shown in Fig. \ref{fig_WS}B. For these calculation the maximum and minimum tensions of the cable are set between 0.5N and 60N; the former to prevent tendon slackness, whereas the latter to prevent tendon failure. The tendons used in this system are 0.19mm diameter UHMW-PE spectra wires (PowerPro, Shimano, Inc., Japan), with a rated strength of 13kgf.

%% file: body/actScaffold.tex
	As mentioned in (\ref{eq_first})-(\ref{eq_last}), deployment of the mechanism depends on angles $\theta_1$ and $\theta_2$. In practice, however, lateral forces on the scaffold will cause high moments in the joints. Combined with the spatial limitations around the joint, it is more feasible to vary length $ a $. As this results in the system being underconstraint, mechanical stops on minimum and maximum angles of $\theta_1$ and $\theta_2$ are required to make the deployed mechanism fully constraint. From \textit{ex vivo} experience we have noticed that the colon also will act as a constraint, enabling smaller deployment diameters in which the system can manipulate the instruments. Initially using the colon as a constraint seems like a disadvantage, however, such a underconstraint mechanism will distribute the forces evenly over the colon length and therefore actually might be beneficial compared to fully constraint mechanisms. This, however, has to be evaluated further in terms of safety during deployment and tendon-actuation. A fully constrained mechanism can be easily implemented by adding a thin link parallel to either $g_1$ or $g_2$, and therefore resulting in $\theta_1 = \theta_2$.
		
	The actuation of length $ a$ is controlled by rotation of a threaded rod, in combination with integrated nut into the device. One end of the threaded rod is constrained in all directions except for rotation, and this rotation therefore results in translation of the nut. In total, four threaded rods are used. The rotation for all threaded rods is actuated by a single externally placed stepper motor. The rotational motion is conveyed with a 1.5 meter torsionally stiff pianowire inside a PTFE outertubing.

%% file: body/benchtop.tex
	\subsubsection{Control Accuracy during user task}
\begin{figure}[b]
	\centering
	\includegraphics[width=3.3in]{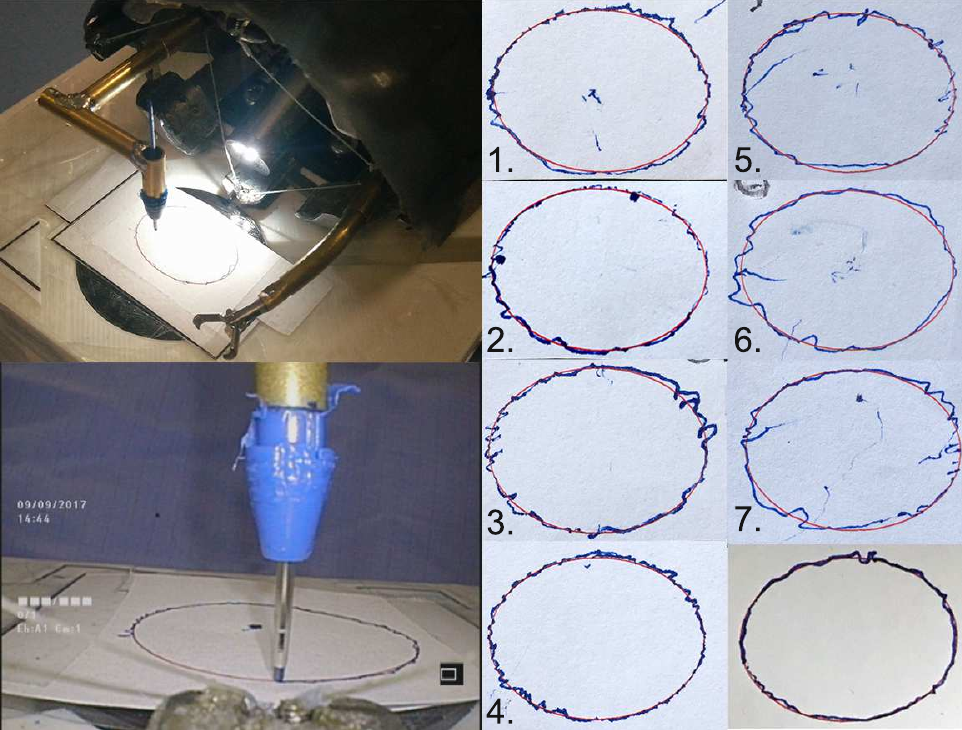}
	\caption{ \textit{Left:} The 15x20mm ellipse as seen during the task from the outside and endoscopic view. The pen is placed perpendicular to the paper with the tip placed at the same position as the curved instrument. \textit{Right:} The tracing task performed by the 7 subjects. The bottom right is the same task performed by the da Vinci robot, by subject 4.}
	\label{fig_mozaik}
\end{figure}

A user study was performed to assess the accuracy of the system. Six engineers and one clinician performed an elliptical tracing task on paper with a pen mounted at the end-effector. To ensure good contact of the pen with the paper, the pen was placed perpendicular, requiring a different end-effector (Fig. \ref{fig_mozaik}). Whereas the shape of the instrument is different, the position of the end-effector is the same as the single-curved overtubes used for the CYCLOPS. The two principle axes of the ellipse are 20mm and 15mm. To assess the accuracy of the system independent from the visualisation used, a high definition 3D flexible endoscope was used (EndoEYE Flex 3D, Olympus, Japan).

The traced tasks, shown in Fig. \ref{fig_mozaik}, have been analysed using the open source scientific image analysis software ImageJ. The total area between the task and the traced line, the total length of the drawn line, and the proportions of the ellipse where no contact of the pen was found, are shown in Table \ref{table_accuracy}. As the error area is the integrated error over the circumferential length of the ellipse, the mean mm deviation per subject can be found by dividing the measured area error by the traced length (excluding areas not traced). As a reference, the best performing subject (no. 4) performed the same task using the da Vinci robot. 

The results show that users are able to perform the tracing task with high accuracy. The da Vinci system, being a commercially available surgical platform, shows slightly better accuracy during the reference task performed by subject 4. However, knowing the current system is still a prototype, the small gap in performance can be bridged in further developments. This comparison is noteworthy, in particular, when realising the low-costs associated with development of the system.

\begin{table*}[t]
	\renewcommand{\arraystretch}{1.3}
	\caption{Metrics of the users (n = 7), while performing the eliptical tracing task}
	\label{table_accuracy}
	\centering
	\begin{tabular}{r||c|c|c||c |c}
		\hline
		\bfseries & \bfseries Mean & \bfseries Std & \bfseries Range & \bfseries Da Vinci (subject 4) &  \bfseries CYCLOPS (subject 4)\\
		\hline\hline	
		Area Error            & 12.69 mm$^2$ & 1.55mm$^2$ & 7.27 - 16.3mm$^2$ & 6.43 mm$^2$ & 7.27mm$^2$\\
		Average mm error deviation     & 0.217mm	   & 0.06mm & 0.133 - 0.302mm & 0.117mm & 0.133mm\\
		Total Length Drawn                      & 76.3mm       & 6.7mm & 59.3 - 77.9mm & 59.5mm & 64.5mm\\
		Ratio drawn line to circumference 		& 122.3\%	   & 12.1\% & 107.0\% - 140.6\% & 107.4\% & 116.4\%\\
		Elliptical circumference not covered    & 2.14\%       & 1.55\% & 0 - 4.97\% & 0.67\% &0.63 \% \\				
		Time to perform the task                & 80.0sec        & 21.9sec  & 43 - 104sec & 70sec & 93sec\\
		\hline
	\end{tabular}
\end{table*}

\subsubsection{Instrument Force Exertion}

\begin{figure}[b]
	\centering
	\includegraphics[width=3.3in]{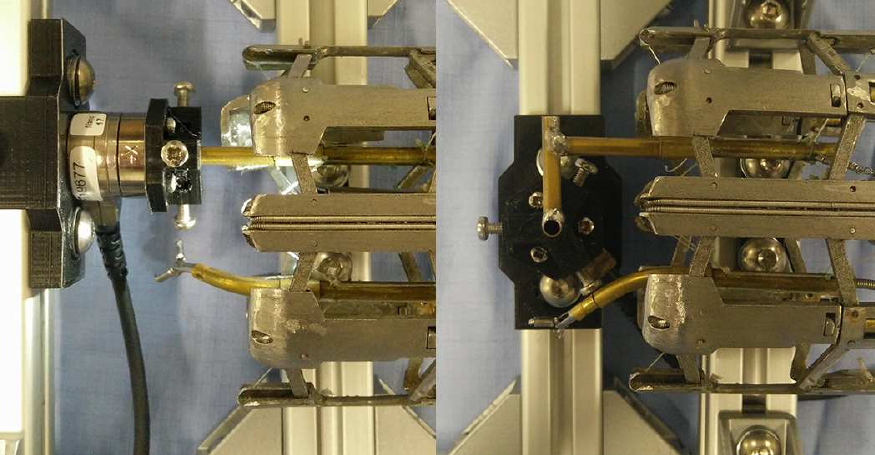}
	\caption{Setup for force measurements on straight tool (left) and angled tool mimicking the curved tool (right).}
	\label{fig_forceSetup}
\end{figure}

The first proof of concept \cite{mylonas2014cyclops} illustrated high force exertion capabilities of up to 65N. The current system has some fundamental differences that require assessment of the forces; these include the use of force transmission cables and different overtube shape and size. Therefore, both are assessed separately. In both cases, a Nano17 6 DOF force transducer is used (ATI Industrial Automoation, Inc., USA) with a 16-bit NI USB-6259 DAQ (National Instruments, USA). The set-ups are shown in Fig. \ref{fig_forceSetup}, The system was set to only control one DOF, while the forces where exerted by slowly moving the end-effector via the master device. The end-effector was placed in the centre of the workspace (homing position). Additionally, the maximum achievable force the system is able to exert in each direction was assessed. For these experiments, the end-effector was placed in one end of the workspace and actuated in the opposite direction (e.g. placing the end-effector in the maximum X+ position, and then pulling in the Y- direction).

\begin{figure}[b]
	\centering
	\includegraphics[width=3.3in]{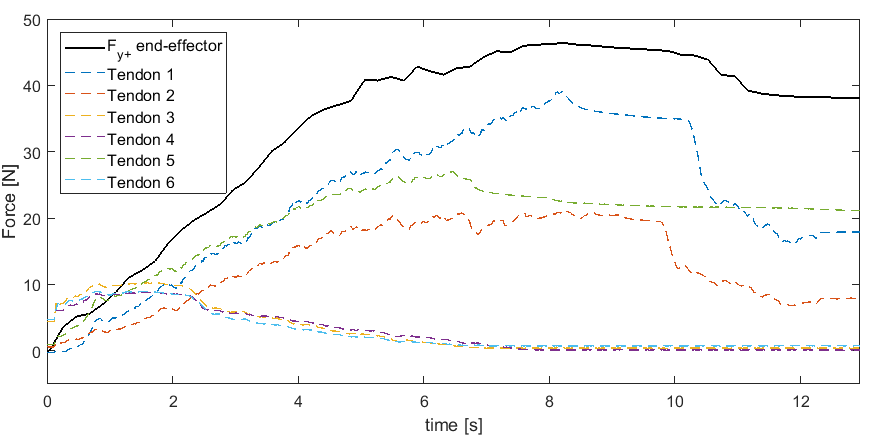}
	\caption{ The measured end-effector forces for the straight instrument. The graph shows that while high forces are achieved, the maximum tension in the cables do not exceed the set maximum force limit of 60N.}
	\label{fig_tendontensions_XposEx}
\end{figure}

The results are shown in Table \ref{table_forces}. During the experiments it became clear that the limitations in forces is not due to the 60N set maximum tension of the cables. This can be seen when looking at the tension of the tendons for the highest force measurement shown in Fig. \ref{fig_tendontensions_XposEx}. The stiffness in the force transmission system (Bowden and tendons) in combination with feed-forward position control, resulted in the system expecting the end-effector to have reached the boundary and thus preventing the motors from continued tensioning. However, despite that an other control method or stiffer transmission system would increase these forces even further, the system is able to exert exceptionally high forces.

\begin{table}
	\renewcommand{\arraystretch}{1.3}
	\caption{End-effector forces, for the straight and curved tool. }
	\label{table_forces}
	\centering
	\begin{tabular}{l||c|c| c}
	
		& \multicolumn{2}{c|}{\bfseries Straight Tool} &  \bfseries Curved Tool\\
		\bfseries &  \textit{Homing} & \textit{Extremity} & \textit{Homing} \\
		\hline\hline	
			\bfseries X+& 21.31 	 N & 46.39 	 N & 19.08 	 N\\
			\bfseries X-& 20.99 	 N & 41.01 	 N & 24.30 	 N\\
			\bfseries Y+& 7.50 	 N & 24.86 	 N & 3.47 	 N\\
			\bfseries Y-& 18.46 	 N & 26.33 	 N & 17.82 	 N\\
			\bfseries Z+& 8.98 	 N & 16.59 	 N & 5.29 	 N\\
			\bfseries Z-& 9.79 	 N & 13.37 	 N & 7.65 	 N\\
		\hline
	\end{tabular}
\end{table}

